\documentclass[onecolumn
]{ceurart}

\usepackage{listings}
\usepackage{mathtools}
\usepackage{todonotes}
\usepackage{hyperref}
\usepackage{cleveref}
\usepackage{csquotes}
\usepackage{caption}
\usepackage{subcaption}

\renewcommand{\thefootnote}{}
\usepackage{graphicx}

\usepackage{pythonhighlight}
\usepackage{algorithm}
\usepackage{algorithmic}
\usepackage{amsmath}
\usepackage{newfloat}
\usepackage{listings}
\floatstyle{ruled}
\newfloat{listing}{tb}{lst}{}
\floatname{listing}{Listing}
\renewcommand{\thefootnote}{\fnsymbol{footnote}}
\begin{document}

\copyrightyear{2022}
\copyrightclause{Copyright for this paper by its authors.
  Use permitted under Creative Commons License Attribution 4.0
  International (CC BY 4.0).}

\conference{Work in process.}


\title{Prompt-based Pre-trained Model for Personality and Interpersonal Reactivity Prediction}

\author[1]{Bin {Li (Corresponding author)}}[email=libincn@hnu.edu.cn,]

\author[2]{Yixuan Weng}[email=wengsyx@gmail.com,]

\address[1]{College of Electrical and Information Engineering, Hunan University}
\address[2]{National Laboratory of Pattern Recognition, Institute of Automation, Chinese Academy Sciences}

\begin{abstract}
This paper describes our proposed method for the Workshop on Computational Approaches to Subjectivity, Sentiment \& Social Media Analysis (WASSA) 2022 shared task on Personality Prediction (PER) and Reactivity Index Prediction (IRI). In this paper, we adopt the prompt-based learning method with the pre-trained language model to accomplish these tasks. Specifically, the prompt is designed to provide knowledge of the extra personalized information for enhancing the pre-trained model. Data augmentation and model ensemble are adopted for obtaining better results. Moreover, we also provided the online software demonstration and the codes of the software for further research.
\end{abstract}

\maketitle
\vskip0.3cm
\par \noindent
\textbf{Code metadata}\\
\vskip0.1cm
\noindent
\begin{tabular}{|l|p{6.5cm}|p{6.5cm}|}
	\hline
	\textbf{Nr.} & \textbf{Code metadata description} & \textbf{Please fill in this column} \\
	\hline
	C1 & Current code version & \textit{v1.0} \\
	\hline
	C2 & Permanent link to code/repository used for this code version & {\url{{https://github.com/WENGSYX/WASSA-ACL-2022}}} \\
	\hline
	C3  & Permanent link to Reproducible Capsule & N/A\\
	\hline
	C4 & Legal Code License   & \textit{MIT License} \\
	\hline
	C5 & Code versioning system used & N/A \\
	\hline
	C6 & Software code languages, tools, and services used & \textit{python,  Django}\\
	\hline
	C7 & Compilation requirements, operating environments \& dependencies & \textit{Python 3, PyTorch, Transformer library, Sqlite3}\\
	\hline
	C8 & If available Link to developer documentation/manual &  \textit{README page}\\
	\hline
	C9 & Support email for questions &N/A \\
	\hline
\end{tabular}\\
\section{Introduction}
Personality can be defined as a set of characteristics (e.g., age, income, and race), which can reflect the differences of individuals in thinking, emotions, and behaviors \cite{vora2020personality}. The strength of personality is worth exploring and pervades human lives everywhere \cite{beck2022mega}. Personality prediction is an interdisciplinary field spanning from psychology to computer science. However, people's personalities can't be directly observed and measured in activity patterns. Humans tend to learn their personalities through language because it is the most prominent way nowadays. To facilitate the personality relative research, we proposed the prompt-based learning method with the pre-trained language model for the Personality Prediction (PER) and Reactivity Index Prediction (IRI) tasks at WASSA@ACL-2022\footnote{https://wassa-workshop.github.io/2022/shared\_task/}, and designed the personality detection software for the online demonstration.
\section{Main Method}
\subsection{Prompt Design}
Prompt Learning \cite{DBLP:journals/corr/abs-2001-07676,DBLP:journals/corr/abs-2107-13586} is considered to be the wise way for providing the pre-trained model with extra knowledge. For this reason, we manually design the prompt to extract relevant knowledge from the pre-trained model for personality prediction, which is presented as the fixed template, i.e., ``A \underline{female}, with \underline{fourth} grade education, \underline{third} race, age is \underline{22} and income is \underline{100000}''.  Specifically, this personalized information is mapped into the tokens, then concatenated with the origin input together in the fixed prompts for learning the joint representation. The codes for the prompt design are presented as follows.
\begin{python}
	text = data['eaasy'] ## data is the input dictionary with several discretized personalized items
	text_prompt = "A {}, with {} grade education is, {} race, age is {}, and income is {}.".format(data['gender'],data['education'],data['race'],data['age'],data['income'])
	
	text = text_prompt + text ## input with prompt
\end{python}
\subsection{Data Augmentation}
Inspired by the work \cite{karimi2021aeda}, we consider the data augmentation with random punctuation marks, i.e., six punctuation marks in \{",", ".", "!", "'", "?"\}. We want to ensure there is at least one inserted mark for more data from one author, so that the model performance will be robust for different noise. The implementation codes are presented as follows.
\begin{python}
	import random
	punctuation = [',','.','!',"'",'?']
	
	for x in data:
	text = x['eaasy']
	for _ in range(20): ## 20 times for data augmentation
	rd = random.randint(0,len(text)-1)
	text[rd].replace(text[rd], text[rd]+random.choice(punctuation))
\end{python}
\subsection{Model Ensemble}
For different pre-trained models \cite{2018bert}, the better choice to improve the final results is to ensemble the pre-trained model \cite{zwanzig1960ensemble}. As a result, we adopt the ensemble method to average the logits for the final prediction. Specifically, we implement the average logits algorithm \cite{inoue2008useful} for the personality and interpersonal reactivity prediction, which can effectively reduce the variance of the logits prediction by averaging the prediction bias produced from different models. The implementation adopts the transformer package\footnote{https://github.com/huggingface/transformers} \cite{wolf-etal-2020-transformers}, where the details are presented below
\begin{python}
	from transformers import DebertaV2Model
	
	model1 = DebertaV2Model.from_pretrained('M1') ## M1, M2 and M3 represent the trained pre-trained models 
	model2 = DebertaV2Model.from_pretrained('M2') 
	model3 = DebertaV2Model.from_pretrained('M3') 
	
	logits1 = model1(**input)['logits']
	logits2 = model2(**input)['logits']
	logits3 = model3(**input)['logits']
	
	logits = (logits1 + logits2 + logits3).mean(1)
\end{python}
{\section{Software for Demonstration}
	Based on websocket technology, real-time communication is realized \cite{ogundeyi2019websocket}. Specifically, we utilized the trained model from the above main method for online software demonstration. The following functions are introduced:
	\begin{enumerate}
		\item Examples of user input can be sent to the back-end server at any time, and real-time calculations are performed through the back-end model.
		\item The server will analyze the input using a cue-based approach and will be able to perform real-time personality information calculations based on pre-trained models.
		\item Finally, the results of the Personality and Reactivity Index are returned through the website with real-time timestamps, where the web page can be presented as the front end.
\end{enumerate}}

\par
\begin{figure}[t]
	\centering
	\includegraphics[scale=0.50]{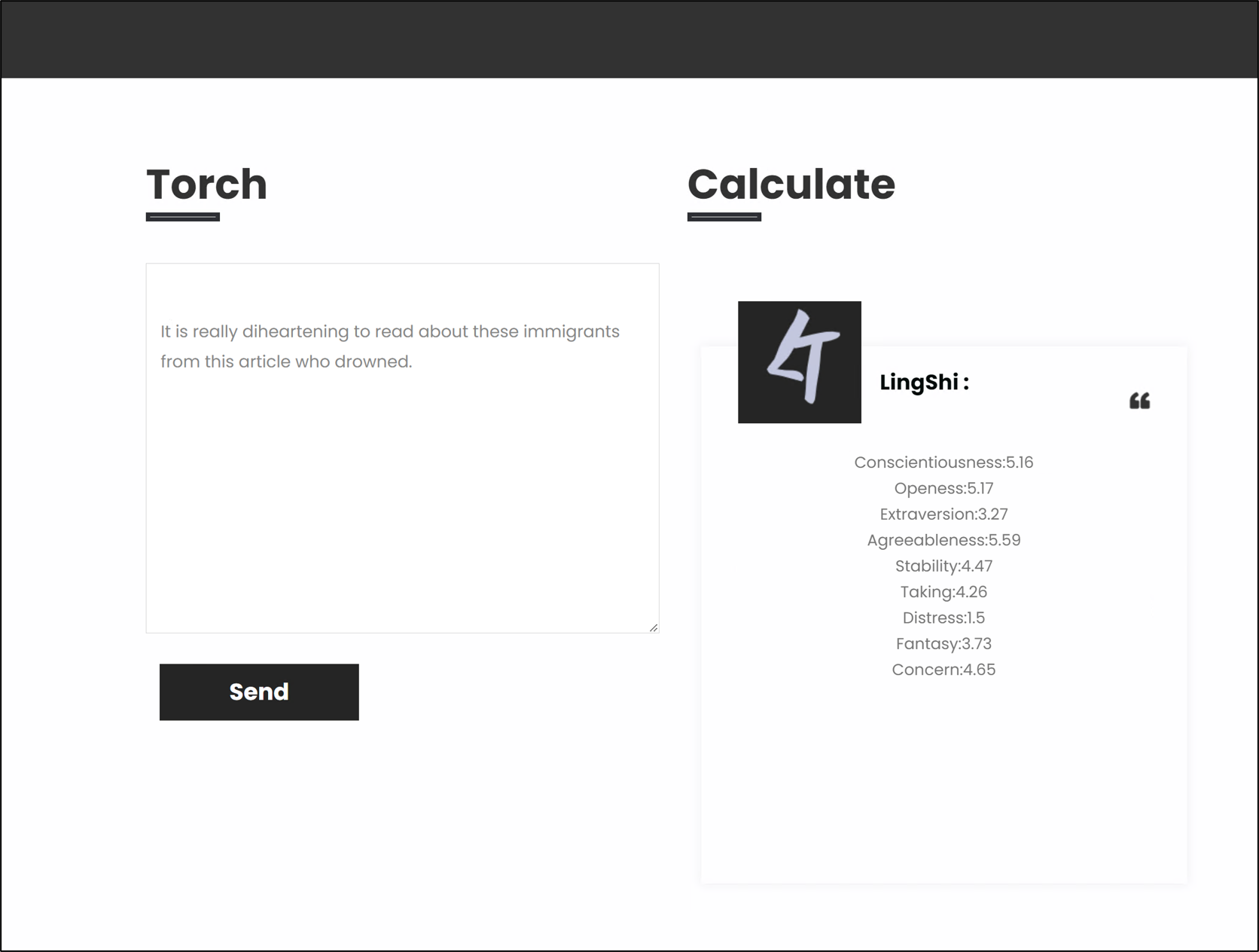}
	\caption{{The demonstration of the proposed software.}} 
	\label{fig3}
\end{figure}
{As shown in Figure \ref{fig3}, we present the software demonstration for the proposed method. We use Django \cite{forcier2008python} to build the entire page and automate subcontracting. Among them, the back-end codes are implemented based on python, and the front-end is HTML and CSS \cite{duckett2011html}.  Based on our implementation, the personality prediction demonstration can be easily realized \cite{junior2019first}. Once deployed the project successfully, just enter the following code to run the software.
	\begin{python}
		python manage.py runserver 0.0.0.0:8000
	\end{python}
	\par
	The online software demonstration is presented at the website \url{http://med.wengsyx.com/LingShi/}, where the codes of the software are open-sourced at \url{https://github.com/WENGSYX/WASSA-ACL-2022}.
	\section{Current Limitations}
	Our system has limitations in two parts. The first is the black-box feature of the pre-trained language model, which makes it hard for us to analyze the interpretability of the personality and reactivity index prediction results. Secondly, our proposed method requires a large number of neurons to be spliced and calculated, so it is difficult to be migrated to the offline mobile scenes.
	\section{Conclusion and Future Improvements}
	In this work, we present our software for personality classification and reactivity prediction, hoping to facilitate the development of this research field with this task. The software devives from an academic competition where we lack sufficient training data from the real world due to privacy constraints. The prompt-based method consists of a pre-trained model with a front-end and back-end presentation framework. In the future, we will focus on more effective prompt designing for performing the personality and interpersonal reactivity prediction.}
\section{Publications and Impacts}
The day-to-day practice of empirical sentiment analysis research can be supported by shifting the focus from engineering \texttt{pytorch} \cite{NEURIPS2019_bdbca288} modules to personalizing predictive design spaces and investigating methods to automatically operate in these spaces.
Our software is implemented based on Django, and the long-term operation of the system can be ensured through the front-end and back-end code erected.
In the future, we will focus on more effective prompt designing for performing the personality and interpersonal reactivity prediction.
\section{Declaration of Competing Interest}
The authors declare that they have no known competing financial
interests or personal relationships that could have appeared to
influence the work reported in this paper.


\bibliography{bibliography}

\appendix

\end{document}